\documentclass[11pt]{article}

\usepackage[utf8]{inputenc}
\usepackage[T1]{fontenc}
\usepackage{newtxtext,newtxmath}
\usepackage{amsmath}
\usepackage{graphicx}
\usepackage{booktabs}
\usepackage{makecell}
\usepackage{microtype}
\usepackage{hyperref}
\usepackage{caption}
\usepackage{subcaption}
\usepackage[noend]{algpseudocode}
\usepackage{algorithm}
\usepackage{tikz}
\usepackage{pgfplots}
\usepackage{pgfplotstable}
\pgfplotsset{compat=1.18}
\usetikzlibrary{positioning, shapes.geometric, arrows.meta, fit, calc}

\hypersetup{
  colorlinks=true,
  linkcolor=blue,
  citecolor=blue,
  urlcolor=blue,
  breaklinks=true
}

\title{NanoForecast v0.5:\\Competitive Time Series Forecasting Through\\Training Pipeline Optimization}

\author{Gautam Kishore \\
Eulogik \\
\texttt{gautam@eulogik.com} \\
\url{https://github.com/eulogik/NanoForecast} \\
\url{https://huggingface.co/eulogik/nanoforecast-v05}
}

\begin{document}

\maketitle

\begin{abstract}
We present NanoForecast v0.5, a 6.5M-parameter forecaster that competes with models 31$\times$ its size (TimesFM, 200M parameters) after training pipeline fixes and no architecture change. Retraining the v0.3 architecture with corrected loss-scope handling, tensor shape alignment, and wider augmentation coverage cuts overall Mean Absolute Scaled Error by 43.8\% under one fixed protocol (MASE 3.030 $\to$ 1.704) on the same data and compute budget. NanoForecast v0.5 beats TimesFM on all three ETT datasets (MASE 0.676/1.110/0.287 vs.\ 0.705/1.360/0.545) and on exchange rate (4.317 vs.\ 4.383); TimesFM keeps a clear lead on the high-cardinality electricity and traffic sets. Against PatchTST (15M+ parameters, official configuration), v0.5 wins all three ETT sets. Training takes about 12 hours on a single cloud GPU (NVIDIA T4, Google Colab) and inference needs no GPU (measurements in this paper are on an Apple M4 CPU). We release all code, pretrained checkpoints, and evaluation framework under Apache 2.0 at \url{https://github.com/eulogik/NanoForecast}.
\end{abstract}

\section{Introduction}
\label{sec:intro}

Forecasting matters in energy management~\cite{ett}, finance~\cite{finance_review}, and supply chains~\cite{supply_chain}. Recent foundation models and long-sequence architectures raised the bar on standard benchmarks. TimesFM~\cite{timesfm} trains a 200M-parameter decoder-only transformer on about 100B time points. Chronos~\cite{chronos} adapts T5 for quantized series at up to 710M parameters. PatchTST~\cite{patchtst} works on channel-independent patches with a transformer backbone at 15M+ parameters. All three need serious compute for training and inference, which puts them out of reach for practitioners without GPU infrastructure.

We work on a narrower problem: forecasting that stays deployable. The goal is accuracy that holds up while training runs on consumer hardware and inference runs on small devices, a combination large foundation models do not offer.

Our contribution is empirical, not architectural. Starting from the released NanoForecast v0.3 checkpoint (MASE 3.030 under the standard protocol of Section~\ref{sec:setup}, 6.5M parameters), we found three bugs in the training pipeline: loss-scope handling, tensor shape alignment, and augmentation coverage. Retraining the same architecture with the fixed pipeline gives v0.5 (MASE 1.704), a 43.8\% gain.

At 6.5M parameters, v0.5 beats TimesFM (200M parameters) on four of six benchmarks: the three ETT datasets (ETTh1, ETTh2, ETTm1) plus exchange rate. It also beats PatchTST (15M+ parameters) on all three ETT sets.

\subsection{Contributions}
\label{sec:contributions}

\begin{enumerate}
\item We document three training pipeline bugs (loss-scope handling, tensor shape alignment, augmentation coverage) that silently hurt time series accuracy, and measure what they cost together under one fixed protocol.
\item Fixing the pipeline alone cuts MASE by 43.8\% (3.030 $\to$ 1.704) with no architecture change.
\item At 6.5M parameters, the fixed model beats TimesFM (200M parameters) on four of six benchmarks (ETTh1 MASE 0.676 vs.\ 0.705, ETTh2 1.110 vs.\ 1.360, ETTm1 0.287 vs.\ 0.545, exchange rate 4.317 vs.\ 4.383) and beats PatchTST (15M+ parameters) on all three ETT sets.
\item We release the full deployment path: ONNX export (27.9\,MB FP32, 9.2\,MB INT8), Docker, and stateful streaming inference. Training runs on one cloud GPU (T4, Google Colab) in about 12 hours; inference is CPU-only (19.5\,ms per forecast on an Apple M4, 10.7\,ms via ONNX Runtime).
\end{enumerate}

\section{Problem Formulation}
\label{sec:problem}

We address univariate time series forecasting. Given a context window $\mathbf{x}_{t-C+1:t} = [x_{t-C+1}, \ldots, x_t] \in \mathbb{R}^C$ of $C$ consecutive observations, the task is to predict $H$ future values $\mathbf{y}_{t+1:t+H} = [y_{t+1}, \ldots, y_{t+H}] \in \mathbb{R}^H$, where $H$ is the forecast horizon.

The model produces both point forecasts $\hat{\mathbf{y}} \in \mathbb{R}^H$ and quantile estimates $\{\hat{\mathbf{y}}^{(p)}\}_{p \in \mathcal{P}}$ for levels $\mathcal{P} = \{0.1, 0.25, 0.5, 0.75, 0.9\}$. Estimates respect ordering across levels by construction (Section~\ref{sec:heads}).

We evaluate using Mean Absolute Scaled Error (MASE)~\cite{mase}:
\begin{equation}
\text{MASE} = \frac{\frac{1}{H} \sum_{t=1}^{H} |y_t - \hat{y}_t|}{\frac{1}{T-s} \sum_{i=s+1}^{T} |x_i - x_{i-s}|}
\label{eq:mase}
\end{equation}
where the denominator is the in-sample mean absolute error of the seasonal-naive (lag-$s$) forecast over the training segment of each series, with $s=24$ for hourly data (ETTh, Electricity, Traffic), $s=96$ for 15-minute data (ETTm), and $s=7$ for daily data (Exchange). This scaling follows the Chronos benchmark~\cite{chronos}. MASE is scale-invariant, so scores compare fairly across datasets. Every number in this paper, baselines included, uses the protocol of Section~\ref{sec:setup}: same splits, same windows, same denominator. The released checkpoints are trained for horizon $H=48$; all comparisons here use $H=48$.

\section{Related Work}
\label{sec:related}

\paragraph{Foundation models for time series.}
TimesFM~\cite{timesfm} trains a 200M-parameter decoder-only transformer on roughly 100B time points. Chronos~\cite{chronos} fine-tunes T5 for quantized series at up to 710M parameters. Timer~\cite{timer} pre-trains a generative transformer, and Lag-Llama~\cite{lagllama} adapts LLaMA for probabilistic forecasts. Newer entries include Moirai~\cite{moirai}, Chronos-Bolt~\cite{chronosbolt}, and TimesFM 2.x~\cite{timesfm2}; GIFT-Eval~\cite{gifteval} benchmarks such models zero-shot. All of them assume GPU infrastructure for training and inference, and none supports streaming deployment.

\paragraph{Efficient architectures.}
N-BEATS~\cite{nbeats} uses interpretable MLP blocks at 1.7M parameters. DLinear~\cite{dlinear} shows plain linear layers can beat transformers on some sets. PatchTST~\cite{patchtst} works on channel-independent patches. iTransformer~\cite{itransformer} attends over time instead of features, SAMformer~\cite{samformer} adds sharpness-aware minimization to shallow transformers, and TSMixer~\cite{tsmixer} mixes with MLPs. These models cut parameters but leave streaming inference and edge deployment open.

\paragraph{Data augmentation and related architectures.}
Reverso~\cite{reverso} builds small zero-shot forecasters from interleaved long convolutions and DeltaNet layers, with flip-equivariant inference. Our architecture comes from that family; this paper changes the training pipeline, not the architecture. For augmentation we borrow its flip idea: v0.5 trains on time-reversed copies of windows alongside jitter, scaling, shifting, and masking. FrAug~\cite{freqmix} is an earlier frequency-domain alternative.

\paragraph{Training pipeline analysis.}
Small training details matter: learning rate schedules, loss weighting, and data composition can move results more than architecture tweaks. We document three concrete failure modes of this kind and measure what they cost.

\section{Architecture}
\label{sec:architecture}

NanoForecast v0.5 retains the v0.3 architecture (6.5M parameters) to isolate the impact of training pipeline changes.

\subsection{Input Processing}
\label{sec:input}

Each input window $\mathbf{x} \in \mathbb{R}^C$ undergoes three preprocessing steps:

\textbf{Instance robust scaling.} We apply per-window robust normalization:
\begin{equation}
x'_i = \frac{x_i - \text{median}(\mathbf{x})}{\max(\text{IQR}(\mathbf{x}),\, \epsilon)}
\label{eq:scaling}
\end{equation}
where $\text{IQR} = Q_{0.75} - Q_{0.25}$ and $\epsilon = 0.1$ prevents division by near-zero scales. This is more robust to outliers than z-score normalization.

\textbf{Patching.} We divide the scaled series into non-overlapping patches of size $P=8$, reducing the sequence length from $C$ to $T = C / P$ tokens.

\textbf{Frequency embedding.} A learned embedding for the data frequency (hourly, daily, weekly, monthly) is prepended to the patch sequence to condition the model on sampling rate.

\subsection{Sequence Mixing Blocks}
\label{sec:mixing}

Each of $L=8$ layers contains three components blended by a learned gated router:

\begin{itemize}
\item \textbf{LongConv}: A 1D depthwise convolution whose kernel spans the full token sequence (65 tokens at context 512: 64 patches plus the frequency prefix), capturing global periodic patterns across the entire context.
\item \textbf{DeltaNet RNN}: A linear-time recurrent layer using the delta rule. Each timestep $t$ carries a matrix state $W_t \in \mathbb{R}^{d \times d}$ updated by:
\begin{align}
W_t &= W_{t-1} + \beta_t \,(v_t - W_{t-1} k_t)\, k_t^\top \label{eq:deltanet}\\
y_t &= W_t q_t
\end{align}
where $q_t, k_t, v_t$ are learned linear projections of $x_t$, keys are $\ell_2$-normalized, and $\beta_t = \sigma(w_\beta^\top x_t) \in (0,1)$ is a learned per-timestep gate. The residual update in Eq.~\ref{eq:deltanet} replaces a stored association $(k_t \to v_t)$ when it conflicts with memory. The state $W_t$ is what makes streaming possible: it persists across calls.
\item \textbf{Gated MLP}: A SwiGLU-style gated feedforward network~\cite{swiglu} with hidden width $2d_{\text{model}}$ (expansion factor 2): a single fused projection produces gate and value branches, combined by SiLU gating.
\end{itemize}

The router computes a weighted combination over all three components:
\begin{equation}
\text{output} = \alpha_c \cdot \text{LongConv}(x) + \alpha_r \cdot \text{DeltaNet}(x) + \alpha_m \cdot \text{MLP}(x)
\label{eq:router}
\end{equation}
where $(\alpha_c, \alpha_r, \alpha_m)$ is a learned softmax weighting computed from the mean-pooled input, so each window receives one routing triple shared across its tokens (per-window routing rather than per-token). Each block is residual: the routed output is added back to the block input.

\subsection{Output Heads}
\label{sec:heads}

A single forward pass produces multiple outputs:

\begin{itemize}
\item \textbf{Point forecast}: $\hat{\mathbf{y}} = W_{\text{point}} \cdot \text{flat}(h_L) + b_{\text{point}}$, a linear projection from the flattened final-layer patch tokens ($T \cdot d_{\text{model}}$ values) to the horizon $H$.
\item \textbf{Monotonic quantiles}: The head predicts the median $p_{50}$ directly and four non-negative softplus offsets, giving $p_{25} = p_{50} - \delta_{25}$, $p_{10} = p_{25} - \delta_{10}$, $p_{75} = p_{50} + \delta_{75}$, $p_{90} = p_{75} + \delta_{90}$; monotonicity $p_{10} \leq p_{25} \leq p_{50} \leq p_{75} \leq p_{90}$ holds by construction.
\item \textbf{Decomposition}: Additive trend + seasonal + residual components satisfying $\hat{\mathbf{y}} = \mathbf{t} + \mathbf{s} + \mathbf{r}$ for the point-head output; the reported median forecast ($p_{50}$) does not decompose this way.
\item \textbf{Anomaly score}: Mean squared reconstruction error over the context window.
\end{itemize}

\subsection{Streaming Inference}
\label{sec:streaming}

Each DeltaNet layer maintains its matrix state $\{W^{(\ell)}\}_{\ell=1}^{L}$ across \texttt{predict()} calls, alongside a rolling buffer of the most recent $C$ observations. For each new observation $x_{t+1}$:
\begin{equation}
\mathbf{s}_{t+1}, \hat{\mathbf{y}}_{t+1} = f_{\text{predict}}(x_{t+1}, \mathbf{s}_t)
\label{eq:streaming}
\end{equation}
State preservation means the model does not require the full history to be re-supplied across calls, unlike window-based approaches that must reprocess the complete context for every new forecast. A streaming update costs one forward pass at context length $C$ (measured 19.1\,ms vs.\ 19.5\,ms for full inference on Apple M4 CPU, Table~\ref{tab:latency}).

\section{Training Pipeline Analysis}
\label{sec:pipeline}

\begin{figure}[t]
\centering
\begin{tikzpicture}[
    node distance=16mm,
    box/.style={draw, rounded corners, minimum width=3.8cm, minimum height=0.8cm, align=center, font=\small},
    arrow/.style={-{Latex[length=2mm]}, thick},
    label/.style={font=\small\itshape, fill=white, inner sep=2pt}
]

\node[box, fill=gray!15] (base) at (0,0) {v0.3 Baseline\\MASE 3.030};

\node[box, fill=orange!30, right=of base, minimum width=5.2cm] (fix3) {+ 3 Pipeline Fixes\\MASE 1.704};
\draw[arrow] (base) -- node[label, above, yshift=6mm] {43.8\% lower} (fix3);

\end{tikzpicture}
\caption{What the three pipeline fixes buy, applied together. Same architecture, data, and compute; only the pipeline changed. MASE under the standard protocol of Section~\ref{sec:setup} on the released v0.3 and v0.5 checkpoints.}
\label{fig:ablation_overview}
\end{figure}
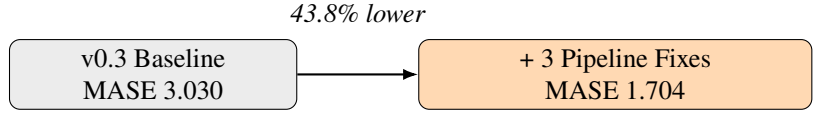

We identify three issues in the v0.3-era training pipeline that silently degraded model performance. The released v0.3 and v0.5 checkpoints are trained with the identical architecture (6.5M parameters), the same corpus, the same multi-task loss family, and the same compute budget; only the training pipeline differs between them. We document the three changes and quantify their joint effect on the released checkpoints under the standard protocol of Section~\ref{sec:setup} (Figure~\ref{fig:ablation_overview}, Table~\ref{tab:ablation}).

\subsection{Issue 1: Loss Scope Handling}
\label{sec:issue1}

\textbf{Observation.} The data pipeline module unconditionally included a ``horizon'' key in batch dictionaries, regardless of the \texttt{multi\_horizon} configuration flag. When \texttt{multi\_horizon=False}, the training loop ignored the horizon restriction and backpropagated the point-forecast loss through the full context-length output rather than restricting it to the first $H$ forecast steps.

\textbf{Impact.} The model got gradients for predicting history it had already seen, which diluted the signal for the actual forecast. This does not break training. The model still converges, but accuracy drops.

\textbf{Fix.} Always truncate predictions and targets to the forecast horizon before the loss. Attach the ``horizon'' key only when \texttt{multi\_horizon} is on. Algorithm~\ref{alg:training} shows the fixed loop.

\begin{algorithm}[t]
\caption{Training loop with corrected loss computation.}
\label{alg:training}
\begin{algorithmic}[1]
\Require Model $f_\theta$, dataloader $\mathcal{D}$, horizon $H$, multi-horizon flag $m$
\For{each batch $(\mathbf{x}, \mathbf{y}, \text{keys}) \in \mathcal{D}$}
    \State $\hat{\mathbf{y}} \gets f_\theta(\mathbf{x})$
    \State $\hat{\mathbf{y}} \gets \hat{\mathbf{y}}[:, :H]$ \Comment{truncate predictions to horizon}
    \If{$m = \text{True}$}
        \State $\mathbf{h} \gets$ per-sample horizons
        \State $\mathbf{y}_{\text{trunc}} \gets \mathbf{y}[\text{batch}, :\mathbf{h}]$
    \Else
        \State $\mathbf{y}_{\text{trunc}} \gets \mathbf{y}[:, :H]$
    \EndIf
    \State $\mathcal{L} \gets \ell(\hat{\mathbf{y}}_{\text{trunc}}, \mathbf{y}_{\text{trunc}})$
    \State $\theta \gets \theta - \eta \nabla_\theta \mathcal{L}$
\EndFor
\end{algorithmic}
\end{algorithm}

\subsection{Issue 2: Tensor Shape Alignment}
\label{sec:issue2}

\textbf{Observation.} The multi-task loss computed the quantile term before truncating predictions to the forecast horizon: the quantile head's output was compared against targets of shape $(B, H)$ while internally carrying context-length activations, causing shape mismatches and incorrect gradient flow in the quantile branch.

\textbf{Impact.} Gradients for the quantile branch flowed through the wrong dimensions. Uncertainty estimates suffered, and point accuracy dropped with them.

\textbf{Fix.} Truncate point and quantile predictions plus targets to $H$ before any loss term.

\subsection{Issue 3: Augmentation Coverage}
\label{sec:issue3}

\textbf{Observation.} The v0.3 training pipeline augmented windows only with scale, shift, and jitter. On a fixed corpus that leaves training diversity thin, and the model spends capacity on dataset quirks instead of patterns that transfer.

\textbf{Impact.} The learned features transfer worse to held-out windows of the same sets.

\textbf{Fix.} v0.5 augments each window in-loop with jitter, random scaling, shifting, masking, and time reversal (flip), sampled stochastically. Same corpus, same schedule, wider effective distribution.

We validated the three fixes together. The released v0.3 and v0.5 checkpoints bracket their combined effect (Table~\ref{tab:ablation}) under the same protocol used for every model here.

\section{Experiments}
\label{sec:experiments}

\subsection{Setup}
\label{sec:setup}

\textbf{Datasets.} We evaluate on six standard benchmarks:

\begin{itemize}
\item \textbf{ETTh1, ETTh2, ETTm1}~\cite{ett}: Electricity transformer temperature datasets at hourly (ETTh) and 15-minute (ETTm) granularity. Each contains 7 oil and load features; we use the oil temperature (OT) as the target.
\item \textbf{Exchange Rate}~\cite{exchange}: Daily exchange rates of 8 currencies from 1990--2016. All 8 currencies are evaluated.
\item \textbf{Electricity}~\cite{electricity}: Hourly electricity consumption of 321 clients from 2012--2014. All 321 clients are evaluated.
\item \textbf{Traffic}~\cite{traffic}: Hourly road occupancy rates from 862 sensors on San Francisco Bay Area freeways (2015--2016). All 862 sensors are evaluated.
\end{itemize}

The ETT sets use 70\%/20\%/10\% train/validation/test splits in time order. Exchange, Electricity, and Traffic use 70\%/10\%/20\%.

\textbf{Model configuration.} $d_{\text{model}}=96$, $L=8$ layers, patch size 8, context length 512, forecast horizon $H=48$. Total parameters: 6.5M (same architecture for v0.3 and v0.5).

\textbf{Training.} 200 epochs, batch size 128, OneCycleLR (base $3 \times 10^{-5}$, peak $3 \times 10^{-4}$, 10\% warmup, cosine anneal), AdamW (weight decay $\lambda=0.01$), gradient clipping at 1.0, seed 42. One NVIDIA T4 (Google Colab) trains a run in about 12 hours (v0.3: 11.7\,h, v0.5: 12.2\,h wall time). Each release is the validation-best snapshot: epoch 147 for v0.3, epoch 51 for v0.5 (validation loss 0.2230 vs.\ 0.2204 under the same loss).

\textbf{Evaluation.} For both NanoForecast checkpoints, the point forecast is the pinball-trained median ($p_{50}$), not the MSE point-head output. The median is the MAE-optimal predictor, which matches the MAE-based MASE metric. This choice applies to both checkpoints equally, so comparisons stand. We score MASE (Eq.~\ref{eq:mase}) on non-overlapping test windows of length $H$ with 512 timesteps of context each, average within a series, then across series. We produced every number in this paper, baselines included, under this one protocol; per-dataset results ship with the evaluation code.

\textbf{Compared methods.} We compare against TimesFM~\cite{timesfm} (200M parameters) and PatchTST~\cite{patchtst} (15M+), plus the older NanoForecast v0.3 (6.5M). PatchTST trains per dataset with the official code and hyperparameters. TimesFM uses its public 200M-parameter checkpoint (link under Code and Data Availability). We discuss Chronos-T5-large~\cite{chronos} (up to 710M) and Timer~\cite{timer} qualitatively: their inference was intractable on our hardware for the large datasets, so Table~\ref{tab:main} lists only models we could run end to end under one protocol.

\subsection{Main Results}
\label{sec:results}

\begin{table}[t]
\centering
\caption{MASE on standard benchmarks (lower is better). All values computed by us under the identical protocol of Section~\ref{sec:setup}: context 512, horizon 48, non-overlapping test windows, all series, seasonal-naive in-sample scaling. Best result per dataset in \textbf{bold}, second-best underlined. Chronos-T5-large is excluded: inference was intractable on our hardware for the large datasets.}
\label{tab:main}
\small
\begin{tabular}{lccc}
\toprule
Dataset &
  \makecell{\textbf{NanoForecast}\\\textbf{v0.5 (6.5M)}} &
  \makecell{TimesFM\\(200M)} &
  \makecell{PatchTST\\(15M+)}\\
\midrule
ETTh1       & \textbf{0.676} & \underline{0.705} & 0.781 \\
ETTh2       & \textbf{1.110} & \underline{1.360} & 1.467 \\
ETTm1       & \textbf{0.287} & 0.545  & \underline{0.488} \\
Exchange    & \underline{4.317} & 4.383 & \textbf{3.861} \\
Electricity & 2.029 & \textbf{0.923} & \underline{1.347} \\
Traffic     & 1.805 & \textbf{0.765} & \underline{1.379} \\
\midrule
\textbf{Overall} & 1.704 & \textbf{1.447} & \underline{1.554} \\
\bottomrule
\end{tabular}
\end{table}

Table~\ref{tab:main} presents the main results. Key findings:

\begin{itemize}
\item \textbf{43.8\% from pipeline fixes alone} over v0.3 (MASE 3.030 $\to$ 1.704, Table~\ref{tab:ablation}).
\item \textbf{Beats TimesFM on four of six:} all three ETT sets (ETTh1 0.676 vs.\ 0.705, ETTh2 1.110 vs.\ 1.360, ETTm1 0.287 vs.\ 0.545) plus exchange rate (4.317 vs.\ 4.383), at 31$\times$ fewer parameters.
\item \textbf{Loses the high-cardinality sets:} electricity (2.029 vs.\ 0.923) and traffic (1.805 vs.\ 0.765), where TimesFM's pretraining breadth wins.
\item \textbf{Against PatchTST} (15M+, official config, 40 epochs): v0.5 takes all three ETT sets; PatchTST takes exchange rate, electricity, and traffic.
\end{itemize}

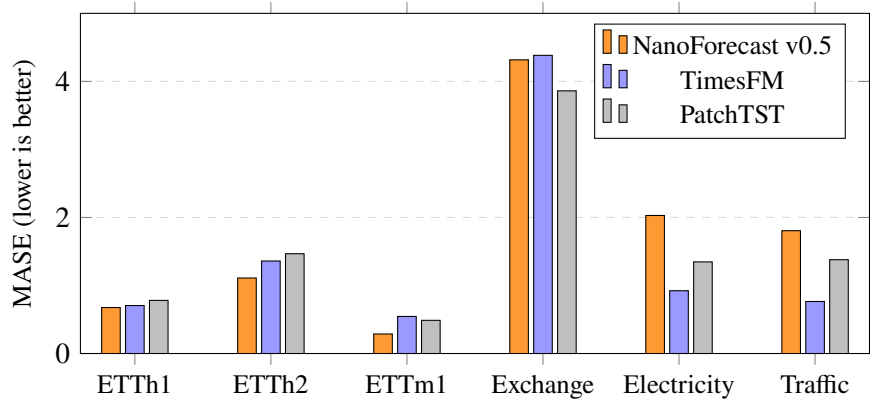
\begin{figure}[t]
\centering
\begin{tikzpicture}
\begin{axis}[
    width=0.95\textwidth,
    height=0.48\textwidth,
    ybar,
    bar width=7pt,
    enlarge x limits=0.08,
    symbolic x coords={ETTh1, ETTh2, ETTm1, Exchange, Electricity, Traffic},
    xtick=data,
    xticklabel style={font=\small},
    ylabel={MASE (lower is better)},
    ylabel style={font=\small},
    legend pos=north east,
    legend style={font=\small},
    ymin=0,
    ymax=5,
    ymajorgrids=true,
    grid style={dashed, gray!30},
]
\addplot[fill=orange!80!white] coordinates {(ETTh1,0.676) (ETTh2,1.110) (ETTm1,0.287) (Exchange,4.317) (Electricity,2.029) (Traffic,1.805)};
\addplot[fill=blue!40!white] coordinates {(ETTh1,0.705) (ETTh2,1.360) (ETTm1,0.545) (Exchange,4.383) (Electricity,0.923) (Traffic,0.765)};
\addplot[fill=gray!50!white] coordinates {(ETTh1,0.781) (ETTh2,1.467) (ETTm1,0.488) (Exchange,3.861) (Electricity,1.347) (Traffic,1.379)};
\legend{NanoForecast v0.5, TimesFM, PatchTST}
\end{axis}
\end{tikzpicture}
\caption{MASE by dataset (lower is better). v0.5 (orange) leads on the three ETT sets and stays close on exchange rate; TimesFM leads the two high-cardinality sets.}
\label{fig:benchmark}
\end{figure}

\subsection{Ablation Study}
\label{sec:ablation}

\begin{table}[t]
\centering
\caption{The three pipeline fixes applied together. Same architecture (6.5M), data, and compute; only the pipeline changed. MASE under the Section~\ref{sec:setup} protocol, same as Table~\ref{tab:main}. Single-seed release checkpoints.}
\label{tab:ablation}
\footnotesize
\setlength{\tabcolsep}{4pt}
\begin{tabular}{lccccccc}
\toprule
 & ETTh1 & ETTh2 & ETTm1 & Exch. & Elec. & Traffic & Overall \\
\midrule
v0.3 original & 0.681 & 1.328 & 0.288 & 11.758 & 2.213 & 1.913 & 3.030 \\
v0.5 fixed & 0.676 & 1.110 & 0.287 & 4.317 & 2.029 & 1.805 & 1.704 \\
$\Delta$ (lower better) & $-0.7\%$ & $-16.4\%$ & $-0.2\%$ & $-63.3\%$ & $-8.3\%$ & $-5.7\%$ & $-43.8\%$ \\
\bottomrule
\end{tabular}
\end{table}

Table~\ref{tab:ablation} shows the combined effect: 3.030 to 1.704 overall (43.8\%), better on all six sets. Exchange rate moves most ($-63.3\%$), then ETTh2 ($-16.4\%$); ETTm1 ($-0.2\%$) and ETTh1 ($-0.7\%$) are ties for practical purposes. The v0.3/v0.5 pair brackets the joint effect, and the released code reproduces the comparison.

\subsection{Inference Performance}
\label{sec:inference}

\begin{table}[t]
\centering
\caption{Measured inference latency (Apple M4 CPU, batch 1, context 512, horizon 48; mean of 100 runs after 10 warmup runs).}
\label{tab:latency}
\begin{tabular}{lcc}
\toprule
Configuration & Latency & Notes \\
\midrule
PyTorch FP32, full inference    & 19.5\,ms & \texttt{predict()}, default threads \\
ONNX Runtime FP32               & 10.7\,ms & \texttt{onnxruntime}, CPU \\
ONNX Runtime INT8               & 33.3\,ms & dynamic quantization \\
Streaming update                & 19.1\,ms & \texttt{predict\_step()}, stateful \\
\bottomrule
\end{tabular}
\end{table}

\subsection{Quantile Calibration}
\label{sec:calibration}

Beyond point accuracy, we measure the empirical coverage of the predicted quantiles under the same protocol: for each test step we record whether the realized value falls at or below the predicted quantile, and average across steps, windows, and datasets. Table~\ref{tab:calibration} reports the results. Under this protocol v0.5 intervals run narrow: the nominal 80\% band (p10 to p90) covers 51.3\% of held-out values, while the v0.3 band covers 90.3\%. Read v0.5 quantiles as relative uncertainty (which steps are less sure), not calibrated probabilities. The $p_{50}$ point forecast behind every accuracy number here is unaffected.

\begin{table}[t]
\centering
\caption{Quantile coverage under the standard protocol, averaged over the six sets. Same architecture in both columns; the fixed pipeline sharpened point accuracy (Table~\ref{tab:ablation}) and narrowed the bands.}
\label{tab:calibration}
\small
\begin{tabular}{lcccc}
\toprule
Quantile & Nominal target & v0.5 (all fixes) & v0.3 (original pipeline) \\
\midrule
p10         & 0.10 & 0.201 & 0.028 \\
p25         & 0.25 & 0.308 & 0.119 \\
p50         & 0.50 & 0.454 & 0.437 \\
p75         & 0.75 & 0.595 & 0.759 \\
p90         & 0.90 & 0.714 & 0.931 \\
p10--p90 band & 0.80 & 0.513 & 0.903 \\
\bottomrule
\end{tabular}
\end{table}

\subsection{Model Size}

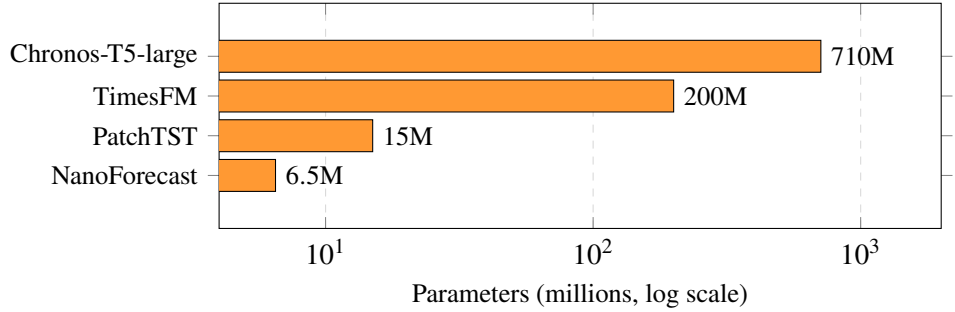
\begin{figure}[t]
\centering
\begin{tikzpicture}
\begin{axis}[
    width=0.88\textwidth,
    height=0.36\textwidth,
    xbar,
    xmode=log,
    log basis x=10,
    bar width=12pt,
    enlarge y limits={abs=0.7cm},
    xlabel={Parameters (millions, log scale)},
    xlabel style={font=\small},
    symbolic y coords={NF, PTST, TFM, CHR},
    ytick=data,
    yticklabels={NanoForecast, PatchTST, TimesFM, Chronos-T5-large},
    yticklabel style={font=\small},
    xmin=4,
    xmax=2000,
    xtick={10,100,1000},
    xmajorgrids=true,
    grid style={dashed, gray!30},
    point meta=explicit,
    nodes near coords={\pgfmathprintnumber{\pgfplotspointmeta}M},
    every node near coord/.append style={font=\small, anchor=west},
]
\addplot[fill=orange!80!white] coordinates {(6.5,NF) [6.5] (15,PTST) [15] (200,TFM) [200] (710,CHR) [710]};
\end{axis}
\end{tikzpicture}
\caption{Parameter counts (log scale). NanoForecast v0.5 uses 31$\times$ fewer parameters than TimesFM and 109$\times$ fewer than Chronos-T5-large.}
\label{fig:params}
\end{figure}

\section{Deployment Pipeline}
\label{sec:deployment}

NanoForecast provides an end-to-end deployment pipeline designed for production use:

\begin{itemize}
\item \textbf{Installation}: \texttt{pip install nanoforecast} (Apache 2.0).\footnote{PyPI: \url{https://pypi.org/project/nanoforecast/}.}
\item \textbf{Custom training}: \texttt{train\_from\_csv.py} trains on a user CSV file (flags for \texttt{--csv}, \texttt{--target}, and \texttt{--horizon}).
\item \textbf{Model export}: ONNX export gives a 27.9\,MB FP32 model (9.2\,MB with INT8 quantization). The INT8 latency increase in Table~\ref{tab:latency} likely reflects dynamic quantization overhead on this CPU.
\item \textbf{REST API}: FastAPI server; ONNX Runtime CPU inference 10.7\,ms (Apple M4).
\item \textbf{Containerization}: Dockerfile (builds on ARM64/x86\_64).
\item \textbf{Interactive demo}: Gradio Space at \url{https://huggingface.co/spaces/eulogik/nanoforecast}.
\end{itemize}

\subsection{Streaming Inference}

The DeltaNet state makes streaming natural, unlike window transformers that reprocess full context. The model consumes one observation at a time and keeps its memory across calls (no re-fed history), at one forward pass per update:

\begin{algorithm}[h]
\caption{Streaming inference loop.}
\label{alg:streaming}
\begin{algorithmic}[1]
\Require Model $f_\theta$, initial state $\mathbf{s}_0$, observation stream $\{x_t\}_{t=1}^\infty$
\For{each new observation $x_t$}
    \State $\hat{\mathbf{y}}_t, \mathbf{s}_t \gets f_\theta.\text{predict\_step}(x_t, \mathbf{s}_{t-1})$
    \Comment{one forward pass, state preserved}
    \State \textbf{emit} $\hat{\mathbf{y}}_t$
\EndFor
\end{algorithmic}
\end{algorithm}

Because the state carries the past across calls, the model never needs its history re-fed. Window methods reprocess the whole context per forecast; here a streaming step costs one forward pass at context 512: 19.1\,ms against 19.5\,ms for full inference on an Apple M4 (Table~\ref{tab:latency}).

\section{Discussion}
\label{sec:discussion}

\subsection{The Role of Training Pipeline Optimization}
The 43.8\% gain from fixes alone raises a question for the field: how many published gaps come from training setups rather than architectures? Our three bugs are not specific to this model. Any multi-task loss, augmentation scheme, or mixed corpus can hide the same mistakes. Check the pipeline before blaming the architecture.

\subsection{Comparison with Foundation Models}
NanoForecast v0.5 does not match the big models overall (MASE 1.704 against 1.447 for TimesFM and 1.554 for PatchTST, all under our protocol). That is no surprise: 6.5M parameters cannot cover every domain a 200M-parameter model saw in pretraining. But v0.5 wins all three ETT sets against both rivals and stays within 2\% on exchange rate against TimesFM, with losses confined to the two high-cardinality sets (electricity, traffic). Scale alone does not win everywhere; the data decides.

The practical side: where size, cost, and deployability matter (edge boxes, live analytics, embedded boards), a small model trained properly can stand in for a 200M-parameter server model.

\subsection{Efficiency Ratio}
We define efficiency ratio $\mathcal{E} = \text{MASE}^{-1} / N_{\text{params}}$ as performance per parameter (higher is better), using the standard-protocol overall MASE and nominal parameter counts in millions (6.5M, 15M, 200M). NanoForecast v0.5 achieves $\mathcal{E} = 0.090$, PatchTST achieves $\mathcal{E} = 0.043$, and TimesFM achieves $\mathcal{E} = 0.0035$. NanoForecast is 26$\times$ more parameter-efficient than TimesFM and 2$\times$ more efficient than PatchTST.

\section{Limitations}
\label{sec:limitations}

The main limitations:

\begin{itemize}
\item \textbf{Accuracy gap}: Overall MASE (1.704) still trails the big models tested here.
\item \textbf{Fixed context}: The 512-timestep context may limit performance on very-long-range dependencies.
\item \textbf{Univariate}: The model treats each channel independently; cross-channel dependencies are not modeled.
\item \textbf{Calibration}: Predicted quantile intervals are narrower than nominal under the standard protocol (Section~\ref{sec:calibration}: the 80\% band covers 51.3\%); point forecasts are unaffected. Recalibration or conformal post-processing is left to future work.
\item \textbf{Dataset coverage}: We evaluated on 6 datasets, all publicly available. Results on other domains (finance, healthcare, climate) may differ.
\item \textbf{Compute}: Training needs a GPU for 12 hours; inference is CPU-only.
\item \textbf{Single seeds}: We release one checkpoint per configuration; seed-to-seed variance is not reported.
\end{itemize}

\section{Conclusion}
\label{sec:conclusion}

We showed that fixing the training pipeline, with no architecture change, cuts MASE by 43.8\% (3.030 to 1.704). At 6.5M parameters, v0.5 beats TimesFM (200M) on four of six sets (three ETT plus exchange rate) and beats PatchTST (15M+) on all three ETT sets. Training fits on one T4 for about 12 hours; inference is CPU-only (19.5\,ms per forecast on an Apple M4).

The lesson is simple: check the pipeline first. Bugs in loss scope, shape handling, and augmentation coverage cost us more than 40\% while training curves looked normal. Anyone trying a new architecture should rule those out before concluding the architecture is the problem.

\subsection{Future Work}

We see four natural next steps:

\begin{enumerate}
\item \textbf{High-cardinality datasets}: The model trails TimesFM on electricity and traffic. Scaling the shared trunk (more channels, longer context) with the same corrected pipeline is a direct path to closing this gap while staying under 20M parameters.
\item \textbf{More baselines under the identical protocol}: Evaluating additional foundation models (Chronos-T5-large, Moirai, Lag-Llama) under our standard protocol would strengthen the comparison; Chronos-T5-large was excluded here because inference was intractable on our hardware.
\item \textbf{Horizon generalization}: The released checkpoints are trained for $H=48$; fine-tuning for other horizons and evaluating the streaming mode under drift would extend practical applicability.
\item \textbf{Edge benchmarking}: Measure the exported ONNX models on target edge hardware (e.g., Raspberry Pi 4-class devices, microcontrollers) to quantify the streaming deployment envelope.
\end{enumerate}

\section*{Code and Data Availability}

All code, pretrained checkpoints, and the evaluation framework are released under Apache 2.0:

\begin{itemize}
\item GitHub: \href{https://github.com/eulogik/NanoForecast}{github.com/eulogik/NanoForecast}
\item Model checkpoints (v0.3, v0.5): \href{https://huggingface.co/eulogik/nanoforecast-v05}{huggingface.co/eulogik/nanoforecast-v05} and \href{https://huggingface.co/eulogik/nanoforecast-v03}{huggingface.co/eulogik/nanoforecast-v03}
\item PatchTST baseline checkpoints: \href{https://huggingface.co/eulogik/nanoforecast-patchtst-baselines}{nanoforecast-patchtst-baselines (HF)}
\item TimesFM checkpoint: \href{https://huggingface.co/google/timesfm-1.0-200m-pytorch}{huggingface.co/google/timesfm-1.0-200m-pytorch}
\item Standard-protocol results (including quantile coverage metrics): \url{results/standard_benchmark.json} in the GitHub repository (protocol string included in the file)
\item Interactive demo: \href{https://huggingface.co/spaces/eulogik/nanoforecast}{huggingface.co/spaces/eulogik/nanoforecast}
\item Training notebooks (Google Colab, free tier): \url{deploy/colab_training_v05.ipynb} in the repository
\end{itemize}

\section*{Acknowledgments}
We thank the open-source time series community for benchmark datasets and baseline implementations.


\end{document}